# Using stigmergy as a computational memory in the design of recurrent neural networks

Federico A. Galatolo, Mario G. C. A. Cimino, and Gigliola Vaglini

*Department of Information Engineering, University of Pisa, 56122 Pisa, Italy*

*{federico.galatolo, mario.cimino, gigliola.vaglini}@ing.unipi.it*

Keywords: Artificial Neural Networks, Recurrent Neural Network, Stigmergy, Deep Learning, Supervised Learning.

Abstract: In this paper, a novel architecture of Recurrent Neural Network (RNN) is designed and experimented. The proposed RNN adopts a computational memory based on the concept of stigmergy. The basic principle of a Stigmergic Memory (SM) is that the activity of deposit/removal of a quantity in the SM stimulates the next activities of deposit/removal. Accordingly, subsequent SM activities tend to reinforce/weaken each other, generating a coherent coordination between the SM activities and the input temporal stimulus. We show that, in a problem of supervised classification, the SM encodes the temporal input in an emergent representational model, by coordinating the deposit, removal and classification activities. This study lays down a basic framework for the derivation of a SM-RNN. A formal ontology of SM is discussed, and the SM-RNN architecture is detailed. To appreciate the computational power of an SM-RNN, comparative NNs have been selected and trained to solve the MNIST handwritten digits recognition benchmark in its two variants: spatial (sequences of bitmap rows) and temporal (sequences of pen strokes).

## 1 INTRODUCTION

Recurrent Neural Networks (RNNs) are today among the most effective solutions for modeling time series, speech, text, audio, video, etc. (Schmidhuber, 2015). An RNN is a special type of NN using its internal state (memory) to process sequences of inputs. This internal memory makes the RNN able to remember the relevant information about the previous samples, in order to model their dynamics.

In contrast to Feed-Forward NN (FFNN), which does not explicitly consider the notion of sequence, in the RNN the input information cycles through a loop. This structure allows the simultaneous processing of both the current and the recent samples. In the RNN, the deep learning algorithm tweaks its weights through gradient descent and backpropagation through time (BPTT, Mazumdar *et al.*, 2008). In essence, BPTT is backpropagation (BP) applied to an equivalent unfolded FFNN. Specifically, Figure 1a shows a basic RNN, made by an MLP layer and a cyclic connection from the output to the input neuron. An RNN with finite response to finite length settles to zero in finite time, and can be modelled as a directed acyclic graph. This RNN can be unfolded and transformed into an FFNN, i.e., an equivalent static MLPs chain, with each MLP working at an instant of time of the finite response, i.e., working without memory (Figure 1b). Thus, within BBTT the error is back-propagated from the last to the first time step. The weights are updated by calculating the error for each time step. Since the unfolded NN is static, it can be trained by BP. However, in case of high number of time steps, the unfolded NN is much larger, and contains a large number of weights, which makes BBTT computationally expensive.

A major issue with BP on large NN chains is related to the gradient descend. In essence, BP goes backwards through the NN to find the partial derivatives of the error with respect to the weights, in order to subtract the error from the weights. Such derivatives are used by the gradient descent algorithm, which iteratively minimizes a given objective function. For better efficiency, the unfolded NN can be transformed into a computational graph of derivatives before training (Goodfellow *et al.*, 2016). A problem when training computational graph is to manage the order of magnitude of gradients throughout a large graph (Pascanu *et al*, 2013). The *exploding gradients* problem occurs when error gradients accumulate during an update. As a result, very large gradients are produced and, in turn, large updates to the network weights of long-term components. This may cause network instability and weights overflow. The problem can be easily solved by clipping gradients when their norm exceeds a given threshold (Goodfellow *et al.*, 2016), by weight regularization, i.e., applying a penalty to the networks loss function for large weight values (Pascanu *et al*, 2013).

On the other side, the *vanishing gradient* problem occurs when the values of a gradient are too small. As a consequence, the model slows down or stops learning. Thus, the range of contextual information that standard RNNs can access is in practice quite limited.

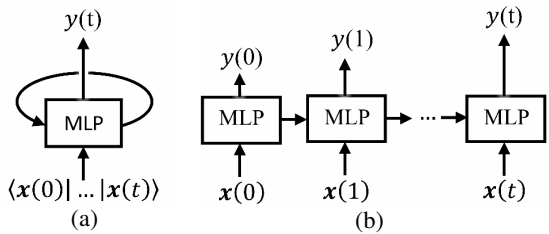

Figure 1: (a) An RNN (b) The equivalent NN unfolded in time.

Long Short-Term Memory (LSTM, Graves *et al.*, 2009) is an RNN specifically designed to address the exploding and vanishing gradient problems. An LSTM hidden layer consists of recurrently connected subnets, called memory blocks. Each block contains a set of internal units, or cells, whose activation is controlled by three multiplicative gates: the input gate, the forget gate, and the output gate. An LSTM network can remember of arbitrary time intervals. The cell decides whether to store (by the input gate), to delete (by the forget gate), or to provide (output gate) information, based on the importance assigned. The assignment of importance happens through weights, which are learned by the algorithm. Since the gates in an LSTM are analog, in the form of sigmoid, the network is differentiable, and trained by BP.

In recent years, LSTM networks have become the state-of-the-art models for many machine learning problems (Greff *et al.*, 2017). This has attracted the interest of researchers on the computational components of LSTM variants.

This paper focuses on a novel concept of computational memory in RNNs, based on *stigmergy*. Stigmergy is defined as an emergent mechanism for self-coordinating actions within complex systems, in which the trace left by a unit's action on some medium stimulates the performance of a subsequent unit's action (Heylighen, 2016). To our knowledge, this is the first study that proposes and lays down a basic design for the derivation of Stigmergic Memory RNN (SM-RNN). In the literature, stigmergy it is a well-known mechanism for swarm intelligence and multi-agent systems. Although its high potential, demonstrated by the use of stigmergy in biological systems at diverse scales, the use of stigmergy for pattern recognition and data classification is still poorly investigated (Heylighen, 2016). As an example, in (Cimino *et al.*, 2015) a stigmergic architecture has been proposed to perform adaptive context-aware aggregation. In (Alfeo *et al.*, 2017) a multi-layer architectures of stigmergic receptive fields for pattern recognition have been experimented for human behavioral analysis. In (Galatolo *et al.*, 2018), the temporal dynamics of stigmergy is applied to weights, bias and activation threshold of a classical neural perceptron, to derive a non-recurrent architecture called Stigmergic NN (S-NN). However, due to the large NN produced by the unfolding process, the S-NN scalability is limited by the vanishing gradient problem. In contrast, the SM-RNN proposed in this paper employs FF-NN as store and forget cells operating on a Multi-mono-dimensional SM, in order to reduce the network complexity.

To appreciate the computational power achieved by SM-RNN, in this paper a conventional FF-NN, an S-NN (Galatolo *et al.*, 2018), an RNN and an LSTM-NN have been trained to solve the MNIST digits recognition benchmark (LeCun *et al.*, 2018). Specifically, two MNIST variants have been considered: spatial, i.e., as sequences of bitmap rows, and temporal, i.e., as sequences of pen strokes (De Jong, E. D., 2018).

The remainder of the paper is organized as follows. Section 2 discusses the architectural design of SM-NNs. Experiments are covered in Section 3. Finally, Section 4 summarizes conclusions and future work.

## 2 ARCHITECTURAL DESIGN

Let us consider, in neuroscience, the phenomenon of selective forgetting that characterizes memory in the brain: information pieces that are no longer reinforced will gradually be lost with respect to recently reinforced ones. This behavior can be modeled by using stigmergy. Figure 2 shows the ontology of an SM, made by four concepts: *Stimulus*, *Deposit*, *Removal,* and *Mark*. In essence, the *Stimulus* is the input of a stigmergic memory. The past dynamics of the *Stimulus* are indirectly propagated and stored in the *Mark*. This propagation is mediated by *Deposit* and *Removal*: *Stimulus* affects *Deposit* and *Removal* which, respectively, reinforces and weakens *Mark*. *Mark* can be reinforced/weakened up to a saturation/finishing level. On the other side, *Mark* itself affects *Deposit* and *Removal*. This behavior can be characterized as recurrent.

Figure 3 shows an example of dynamics of a mono-dimensional SM, i.e., a real-valued *mark* variable, generically called $m(t)$. Specifically, the mark starts from $m(0)$, and for $t = 0,...,4$ it undergoes a weakening by $\Delta m^-(0),...,\Delta m^-(4)$, respectively, up to the finishing level $\underline{m}$. For $t = 11,...,13$ the mark variable undergoes a reinforcement by $\Delta m^+(11), ... , \Delta m^+(13)$, respectively, up to the saturation level $\overline{m}$.

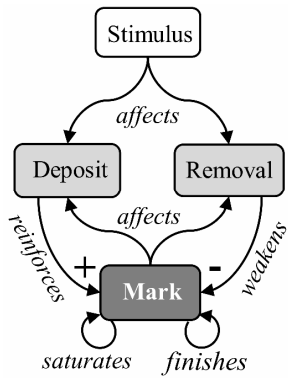

Figure 2: Ontology of a stigmergic memory

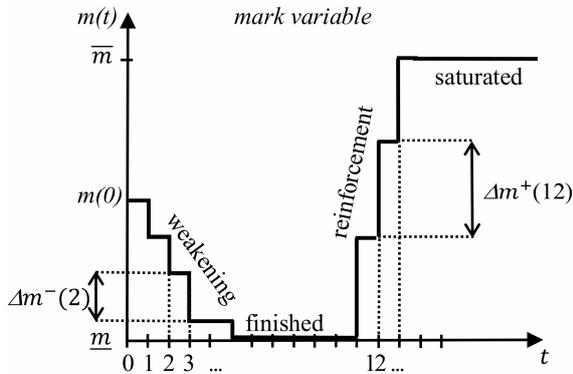

Figure 3: Example of dynamics of a mono-dimensional stigmergic memory.

Let us consider a mono-dimensional *Stimulus*, i.e., a real-valued variable generically called $s(t)$. For each $t$, $\Delta m^+(t)$ and $\Delta m^-(t)$ are determined by *Deposit* and *Removal*, respectively, on the basis of $s(t)$. Thus, $m(t)$ is a sort of aggregated memory of the $s(t)$ dynamics. The relationship between $m(t)$ and $s(t)$ is not prefixed. By using $m(t)$ to feed a subsequent classification or regression unit, this relationship can be trained via supervised learning.

According to this concept, Figure 4 shows the structure of an SM-RNN based classification unit. Here, the Deposit, Removal, and Classification MLPs are realized by spatial FF-MLPs. The SM is based on an array of *M* mono-dimensional mark variables, where *M* is also equal to the number of outputs of the Deposit and Removal MLPs, as well as to the number of inputs of the Classification MLP.

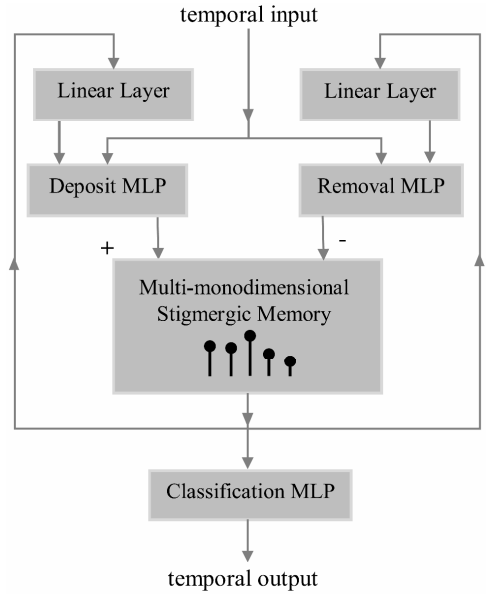

Figure 4: Structure of an SM-RNN based classification unit

Specifically, the Linear Layer at the input of Deposit and Removal MLP is a single layer of linear neurons, i.e., neurons with linear activation function. It performs a linear projection of the SM data.

The three MLPs have the same structure, represented in Figure 5: (i) an Input Linear Layer; (ii) a PReLU (Parametric Rectified Linear Unit) activation function, which solves the vanishing gradient problem; (iii) an Output Linear Layer; (iv) a ReLU (Rectified Linear Unit) activation function, for the Deposit and Removal MLPs, or a PReLU activation function, for the Classification MLP, respectively.

## 3 EXPERIMENTAL STUDIES

The architecture of an SM-NN has been developed with the PyTorch framework (PyTorch, 2018) and made publicly available on GitHub (GitHub, 2018).

To appreciate the computational power of an SM-RNN, different NNs have been trained to solve the MNIST digits recognition benchmark (LeCun *et al.*, 2018): an SM-RNN, an FF-NN, an S-NN, an RNN and an LSTM-NN.

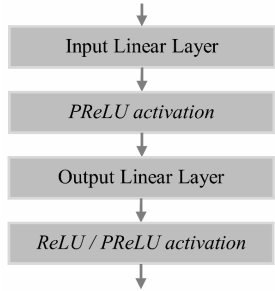

Figure 5: Structure of a Deposit, Removal, or Classification MLP

The purpose is twofold: to measure the computational power of SM-RNN with respect to FF-NN, and to compare the performances of the other existing temporal NN. For this purpose, the following two variants of the MNIST benchmark have been used.

In the *Spatial MNIST* dataset (LeCun et al., 2018), the input image is made by 28×28 = 784 pixels, and the output is made by 10 classes corresponding to decimal digits. In the case of FF-NN, the handwritten character is supplied in the form of full static bitmap. For the other NNs, the handwritten character is supplied row by row, in terms of 28 inputs, over 28 subsequent instants of time. In this case, once provided the last chunk, the NN provides the corresponding output class.

In the *Temporal MNIST* dataset (De Jong, E. D., 2018), the handwritten character is supplied as a sequence of pen strokes. In this case, at each instant of time $t$, the next input is provided as a movement in the horizontal and vertical directions $(dx(t), dy(t))$. Once provided the last pen stroke, the NN provides the corresponding output class. As an example, Figure 6 shows the representation of a handwritten digit in the Spatial MNIST (a) and Temporal MNIST (b).

To adequately compare the different NNs, the following methodology has been used. First, the FF-NN and the SM-RNN have been dimensioned to achieve their best classification performance. Secondly, the S-NN, RNN and LSTM-NN have been dimensioned to have a similar number of parameters with respect to the SM-RNN.

Overall, the data set is made of 70,000 images. At each run, the training set is generated by random extraction of 60,000 images; the remaining 10,000 images makes the testing set.

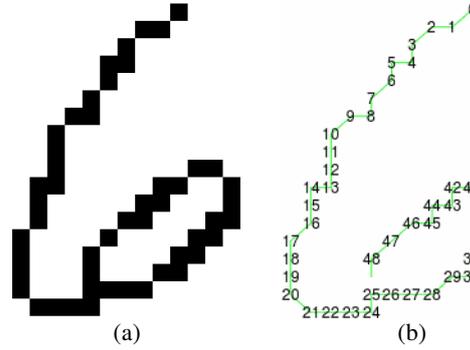

Figure 6: representation of a handwritten digit in the Spatial MNIST (a) and Temporal MNIST (b)

Table 1 shows the overall complexity of each NN. The complexity values correspond to the total number of parameters. Specifically, the SM is made by $M$ = 15 mark variables. Thus, the Deposit and Removal MLPs topology is made by 28 (temporal) + 15 (Linear Layer) inputs, i.e. 43 inputs. The Linear Layer before the Deposit and Removal MLPs contains 15·15 weights + 15 biases = 240 parameters. The Deposit/Removal MLPs contains the Input Linear Layer (43·20 weights + 20 biases = 880 parameters). The PReLU contains 20 parameters. The Output Linear Layer contains 20·15 weights + 15 biases = 315 parameters. The Classification MLP contains the Input Linear Layer (15·10 weights + 10 biases = 160 parameters). The PReLU contains 10 parameters. The Output Linear Layer contains 10·10 weights + 10 biases = 110. Thus, the total number of parameter is 240·2 + (880+20+315) ·2 + (160+10+110) = 3,190.

For a detailed calculation of the complexity of the other NNs, the interested reader is referred to (Galatolo *et al.* 2018).

Table 1: Performance and complexity of different NNs solving the Spatial MNIST digits recognition benchmark.

| Neural Network | Complexity | Classification rate |
|---|---|---|
| SM-RNN | 3,190 | .965 ± 0.056 |
| FF-NN | 328,810 | .951 ± 0.0026 |
| LSTM-RNN | 3,360 | .943 ± 0.011 |
| S-NN | 3,470 | .927 ± 0.016 |
| RNN | 3,482 | .766 ± 0.033 |

In addition, Table 1 shows the performance evaluations, which are based on the 99% confidence interval of the classification rate (i.e., the ratio of correctly classified inputs to the total number of inputs), calculated over 10 runs.

The Adaptive Moment Estimation (Adam) method (Kingma *et al.*, 2015) has been used to compute adaptive learning rates for each parameter of the gradient descent optimization algorithms, carried out with batch method.

It is apparent from the table that the SM-RNN outperforms the other NNs, in terms of both complexity and classification rate. Specifically, the FF-NN employs a very large number of parameters, about two order of magnitude larger with respect to the SM-RNN. To assess the quality of the training process, Figure 7 and Figure 8 show a scenario of classification rate and the function, on the training set, against the number of iterations, respectively. The loss function is calculated as the Negative Log-Likelihood (NLL) using the softmax activation function at the output layer of the NN, which is commonly used in multi-class learning problems.

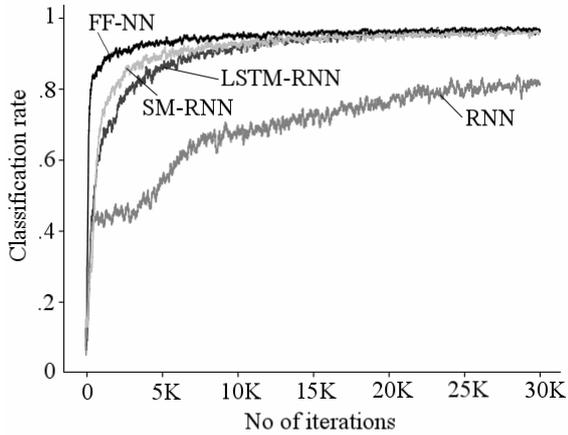

Figure 7: Scenario of classification rate on training set against number of iterations, for the Spatial MNIST data set

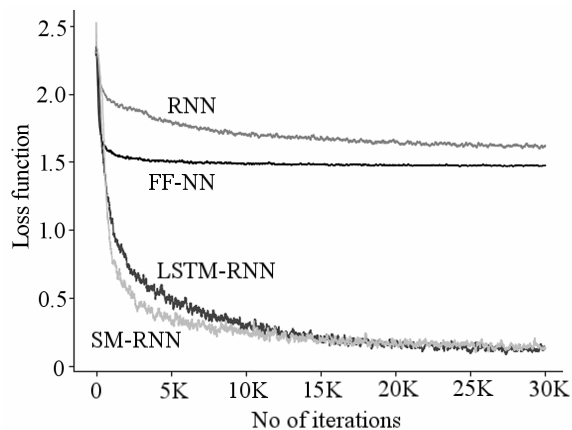

Figure 8: Scenario of loss function on training set against number of iterations, for the Spatial MNIST data set

With regard to the Temporal MNIST data set, three kinds of temporal NNs have been used, i.e., SM-RNN, LSTM-RNN, and RNN.

Table 2 shows the overall complexity of each NN. The complexity values correspond to the total number of parameters. Specifically, the SM is made by $M = 30$ mark variables. Thus, the Deposit and Removal MLPs topology is made by 4 temporal inputs (i.e., the horizontal and vertical directions, the stroke and digit ends) + 30 (Linear Layer) inputs, i.e. 34 inputs. The Linear Layer before the Deposit and Removal MLPs contains 30·30 weights + 30 biases = 930 parameters. The Deposit/Removal MLPs contains the Input Linear Layer (34·20 weights + 20 biases = 700 parameters). The PReLU contains 20 parameters. The Output Linear Layer contains 20·30 weights + 30 biases = 630 parameters. The Classification MLP contains the Input Linear Layer (30·20 weights + 20 biases = 620 parameters). The PReLU contains 20 parameters. The Output Linear Layer contains 20·10 weights + 10 biases = 210. The activation ReLU contains 10 neurons. Thus, the total number of parameters is 930·2 + (700+20+630)·2 + (620+20+ 210+10) = 5,420.

The Recurrent NN is made by the following layers: an Input Linear Layer (34·50 weights + 50 biases = 1,750 parameters), a PReLU (50 parameters), an Output Linear Layer (50·30 weights + 30 biases = 1,530 parameters), and an activation PReLU (30 parameters). Thus, the total number of parameters is 1,750 + 50 + 1,530 + 30 = 3,360. In the Recurrent NN, each output neuron has a backward connection to the input and to the Classification MLP which, in turn, is made by the following layers: the Input Linear Layer (30·50 weights + 50 biases = 1,550 parameters), the PReLU (50 parameters), the Output Linear Layer (50·10 weights + 10 biases = 510), and the activation PReLU (10 neurons). Thus, the total number of parameters of the Recurrent and the Classification NNs is 3,360+1,550+50+510+10 = 5,480.

The LSTM-RNN fed by 4 inputs. For each LSTM layer, the number of parameters is calculated according to the well-known formula $4 \cdot o \cdot (i+o+1)$, where $o$ and $i$ is the number of outputs and inputs, respectively. The topology is made by a 4×20 LSTM layer, a 20×20 LSTM layer, and a 20×10 Output Linear layer. Thus, the overall number of parameters is $4 \cdot 20 \cdot (4+20+1) + 4 \cdot 20 \cdot (20+20+1) + 20 \cdot 10 + 10 = 5,490$.

In addition, Table 2 shows the performance evaluations, based on the same criteria detailed for Table 1. It is apparent from the table that the SM-

RNN and the LSTM-RNN are equivalent in terms of classification rate. In contrast, the RNN is not able to gain a sufficient stability and accuracy. To assess the quality of the training process, Figure 9 and Figure 10 show a scenario of classification rate and loss function, on the training set, against the number of iterations, respectively. The loss function of Figure 10 is calculated as in Figure 8.

Table 2: Performance and complexity of different NNs solving the Temporal MNIST data set.

| Neural Network | Complexity | Classification rate |
|---|---|---|
| SM-RNN | 5,420 | .9467 ± 0.0076 |
| LSTM-RNN | 5,490 | .9496 ± 0.0027 |
| RNN | 5,480 | .7295 ± 0.1101 |

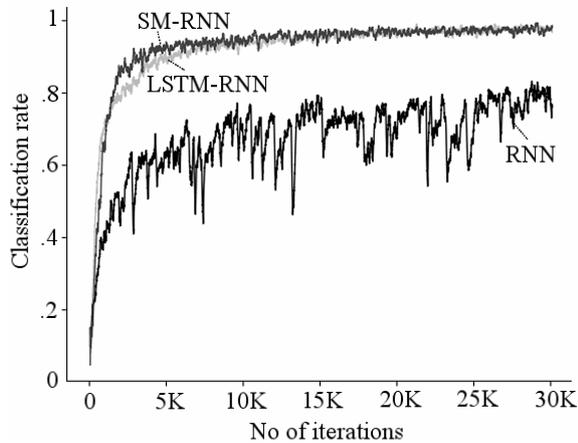

Figure 9: Scenario of classification rate on training set against number of iterations, for the Temporal MNIST data set

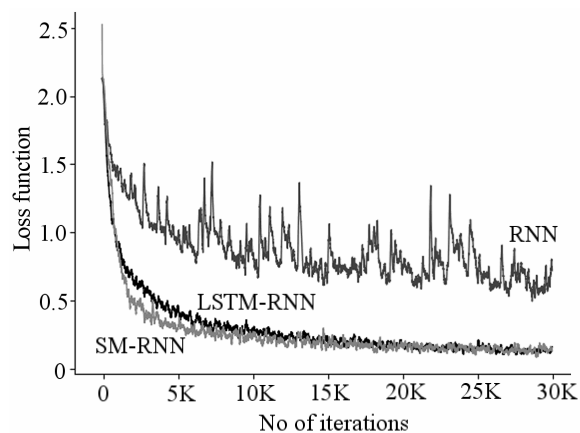

Figure 10: Scenario of loss function on training set against number of iterations, for the Temporal MNIST data set

Overall, the proposed SM-RNN shows a very good convergence with respect to the other NNs. In consideration of the relative scientific maturity of the other comparative NNs, the experimental results with the novel SM-RNN looks very promising, and encourage further investigation activities for future work

## 4 CONCLUSIONS

In this paper, the concept of computational stigmergy is used as a basis for developing a Stigmergic Memory for Recurrent Neural Networks. Some important issues in the research field, related to the gradient descend, are first discussed. The novel architectural design of the SM-RNN is then detailed. Finally, the effectiveness of the approach is shown via experimental studies, carried out on the spatial and temporal MNIST data benchmarks. The SM-RNN can be appreciated for their impressive computational power with respect to the other types of RNNs. Experimental results are promising, and show that the SM-RNN outperforms the FF-NN, the S-NN, the RNN, and is equivalent to LSTM-NN, in terms of both complexity and classification rate. Future work will focus on further experimentation and investigation.

## ACKNOWLEDGEMENTS

This work was partially carried out in the framework of the SCIADRO project, co-funded by the Tuscany Region (Italy) under the Regional Implementation Programme for Underutilized Areas Fund (PAR FAS 2007-2013) and the Research Facilitation Fund (FAR) of the Ministry of Education, University and Research (MIUR).
This research was supported in part by the PRA 2018_81 project entitled "Wearable sensor systems: personalized analysis and data security in healthcare" funded by the University of Pisa